\documentclass[times,final]{elsarticle}

\usepackage{filecontents}

\usepackage{flushend}
\usepackage{framed,multirow}
\usepackage{helvet}
\usepackage{courier}
\usepackage{etoolbox}
\usepackage[usenames,dvipsnames]{xcolor}
\usepackage{tikz}
\usetikzlibrary{shapes.geometric}
\usetikzlibrary{arrows, calc, matrix}
\usepackage{amssymb, amsmath}
\usepackage{amsthm}
\usepackage{latexsym}
\usepackage[ruled,vlined]{algorithm2e}
\usepackage{graphicx}
\usepackage{epsf}
\usepackage{url}
\definecolor{newcolor}{rgb}{.8,.349,.1}

\usepackage{caption}
\usepackage{subcaption}
\usepackage{microtype}
\usepackage[font=small,floatrowsep=qquad,captionskip=5pt]{floatrow}
\usepackage{tabularx}
\usepackage{rotating}
\usepackage{multirow}
\usepackage{xspace}
\usepackage{enumerate}

\usepackage{booktabs}
\usepackage{arydshln}
\usepackage[numbers]{natbib}

\newcolumntype{H}{>{\setbox0=\hbox\bgroup}c<{\egroup}@{}}
\newcommand{\AlgName}{RACE\xspace}
\newcommand{\AlgAbbr}{\underline{Ra}ndom \underline{C}ompr\underline{e}ssion\xspace}

\newcommand{\argmin}{\operatornamewithlimits{argmin}}

\makeatletter
\def\@author#1{\g@addto@macro\elsauthors{\normalsize%
    \def\baselinestretch{1}%
    \upshape\authorsep#1\unskip\textsuperscript{%
      \ifx\@fnmark\@empty\else\unskip\sep\@fnmark\let\sep=,\fi
      \ifx\@corref\@empty\else\unskip\sep\@corref\let\sep=,\fi
      }%
    \def\authorsep{\unskip,\space}%
    \global\let\@fnmark\@empty
    \global\let\@corref\@empty  
    \global\let\sep\@empty}%
    \@eadauthor={#1}
}
\makeatother

\journal{Pattern Recognition Letters}

\begin{document}

\begin{frontmatter}

\title{Online Multi-Label Classification: A Label Compression Method}

\author{Zahra Ahmadi} 
\ead{zaahmadi@uni-mainz.de}
\author{Stefan Kramer}

\address{Institut f\"{u}r Informatik, Johannes Gutenberg-Universit\"{a}t, Staudingerweg 9, Mainz 55128, Germany}

\begin{abstract}
Many modern applications deal with multi-label data, such as functional categorizations of genes, image labeling and text categorization. Classification of such data with a large number of labels and latent dependencies among them is a challenging task, and it becomes even more challenging when the data is received online and in chunks. Many of the current multi-label classification methods require a lot of time and memory, which make them infeasible for practical real-world applications. In this paper, we propose a fast linear label space dimension reduction method that transforms the labels into a reduced encoded space and trains models on the obtained pseudo labels. Additionally, it provides an analytical method to update the decoding matrix which maps the labels into the original space and is used during the test phase. Experimental results show the effectiveness of this approach in terms of running times and the prediction performance over different measures.
\end{abstract}


\end{frontmatter}

\section{Introduction} \label{intro}
Standard classification is the task of assigning the correct class to previously unknown test instances based on training instances. Training data consist of a set of features and an associated target class or class label. Many modern data mining applications, however, need to deal with more than one label per instance. This problem is called multi-label learning. Early studies focused on multi-label text categorization~\cite{mccallum1999}, and gradually the topic attracted attention from different communities, such as bioinformatics~\cite{clare2001knowledge} and image labeling~\cite{boutell2004}. Multi-label classification can be viewed as a generalization of multi-class classification where labels do not exclude each other and may have unknown dependencies among each other as well as with the features. One goal of multi-label classification is thus to take advantage of hidden label correlations to improve classification performance.

Besides multiple interdependent class labels, we are facing a huge increase in the size of data becoming available. In many cases, this data is received as a stream, e.g. a stream of sensor data or an email text stream. Because of the evolving nature of data streams, we cannot record all the instances and cannot assume previous data can be scanned an arbitrary number of times. Therefore, we need efficient methods in terms of time and space complexity. This makes multi-label classification on data streams even more challenging. 

A common approach to multi-label classification is to transform the problem into one or more single-label problems (e.g.~\cite{read2011classifier,tsoumakas2007multi}). The important advantage of problem transformation methods is the possibility of using any available single-label base classifier. This makes these approaches more flexible and generally applicable. However, they may suffer from high computational complexity (as in the Label Powerset method~\cite{tsoumakas2007random}) or ignore the label dependencies (as in the Binary Relevance method~\cite{tsoumakas2007multi}); both are important features in the classification of multi-label streams. One successful type of transformation methods compresses the original label space to a compressed representation~\cite{hsu2009,tai2012,wicker2012,zhou2012}. These methods accelerate the learning process by training fewer binary classifiers on compressed label sets, which makes them suitable for multi-label stream classification. However, all the existing methods need to access the whole training data at once, and none of them has been adjusted for the online setting.

\begin{figure}[tb]
\centering
\caption{General framework of the \AlgName approach.} \label{OLR2}
\begin{tikzpicture}
  
  \filldraw[color={rgb:orange,1;yellow,2;pink,5;white,1}] (-0.5,2.5) [rounded corners=5pt] -- (3.5,2.5) -- (3.5,2.85) -- (-0.5,2.85) -- cycle;
  \node[rectangle,minimum size=4pt,inner sep=0pt,outer sep=0pt] at (1.5,2.67) {$input\quad labels$};
  \filldraw[color={rgb:gray,2;white,2}] (0.1,1.45) [rounded corners=5pt] -- (2.9,1.45) -- (3.5,2.45) -- (-0.5,2.45) -- cycle;
  \node[rectangle,minimum size=4pt,inner sep=0pt,outer sep=0pt] at (1.5,1.95) {$A$};
  \filldraw[color={rgb:orange,4;yellow,1}] (0.1,1.05) [rounded corners=5pt] -- (0.1,1.4) -- (2.9,1.4) -- (2.9,1.05) -- cycle;
  \node[rectangle,minimum size=4pt,inner sep=0pt,outer sep=0pt] at (1.5,1.22) {$pseudo\; labels$};
  \filldraw[color={rgb:gray,2;white,2}] (-0.5,0) [rounded corners=5pt] -- (0.1,1) -- (2.9,1) -- (3.5,0) -- cycle;
  \node[rectangle,minimum size=4pt,inner sep=0pt,outer sep=0pt] at (1.5,0.5) {$\beta$};
  \filldraw[color={rgb:orange,1;yellow,2;pink,5}] (-0.5,-0.05) [rounded corners=5pt] -- (3.5,-0.05) -- (3.5,-0.4) -- (-0.5,-0.4) -- cycle;
  \node[rectangle,minimum size=4pt,inner sep=0pt,outer sep=0pt] at (1.5,-0.22) {$output\quad labels$};

\end{tikzpicture}
\end{figure}
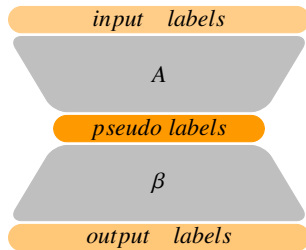

In this paper, we propose a novel online linear label compression approach, called \AlgName (\AlgAbbr). For each new batch of data, we (1) compress the label space into a smaller random space, then (2) update the single-label online classifiers or regressors on this compressed label set (we call them pseudo labels), and in the end (3) update the existing decompressing function analytically with the recent data batch. Figure~\ref{OLR2} represents the general framework of \AlgName.
The main feature of this approach is to process each data batch once and analytically, hence it is not iterative in finding encoding/decoding functions which leads to a faster method compared to other batch label space compression methods in the literature. In addition, it does not have many control parameters to be set; the only parameter is the dimension of the compressed space. Experiments demonstrate that \AlgName performs favorably compared to other multi-label stream classification methods in terms of running times and predictive performance across different datasets.

The remainder of this paper is organized as follows: Section~\ref{related} presents a background on multi-label learning as well as using random features in classification problems. Section~\ref{method} explains the proposed method. Experiments are presented in section~\ref{expr}. Finally, section~\ref{conc} concludes the paper. 

\section{Related Work} \label{related}
Multi-label classification approaches usually fall into two categories: algorithm adaptation and problem transformation methods. Several single-label classification algorithms have been adapted to multi-label data (e.g. ML-kNN~\cite{zhang2005k}). In contrast, problem transformation methods convert the original multi-label problem into one or several single-label problems and apply one of the existing single-label learning methods to the transformed data. One well-known method from this category is called Binary Relevance (BR), which is similar to the one-versus-all approach for multi-class classification. BR assumes the labels are independent and trains a separate model for each label. In many real-world applications, this assumption is not valid. Later, to alleviate this problem, Classifier Chains (CC)~\cite{read2011classifier} were proposed. CC adds links to the classifiers in BR in such a way that the feature space of each link in the chain is extended with the label associations of all previous links. As the order of the chain can influence the performance, an Ensemble of Classifier Chains (ECC) was proposed. A detailed taxonomy of multi-label methods was presented by Gibaja and Ventura \cite{Gibaja2015}. 

With an increase in the number of labels, many of the standard multi-label classification methods that work on the original label space (e.g. BR and ECC) become computationally infeasible. Hence, new strategies for reducing the label space have been presented. Multi-label prediction via Compressed Sensing (CS)~\cite{hsu2009} assumes sparsity in the label set and encodes labels using a small number of linear random projectors. Although the encoding function is linear, the decoder is not. For each test instance, CS needs to solve an optimization problem related to its sparsity assumption. Hence, CS can be time consuming during prediction. 
Unlike CS, the projection matrix in PLST~\cite{tai2012} captures the correlations between labels using Singular Value Decomposition (SVD) of the label matrix. This approach guarantees the minimum encoding error on the training set. Both encoding and decoding functions are computed from the SVD. 
Linear Gaussian random projection is another form of transforming labels~\cite{zhou2012}. The method uses a series of Kullback-Leibler divergence based hypothesis tests for decoding. 
It performs a recursive clustering algorithm and extracts an auxiliary distilled label set of frequently appearing label subsets. However, this step is empirically expensive. 
MLC-BMaD \cite{wicker2012} is a label compression method that uses Boolean Matrix Decomposition (BMaD) to factorize the label matrix into the product of a Boolean code matrix and a Boolean decoding matrix. The method needs all the training data at once to extract the compressed label set. 
Instead of transforming label sets, a label subset selection method based on group-sparse learning was proposed~\cite{BalasubramanianL12}. However, the optimization problem to find the best subset is still computationally expensive. To alleviate this issue, a randomized sampling method based on the Column Subset Selection Problem (CSSP) has been proposed~\cite{BiK13}. 
Lately, deep learning methods have been employed in order to extract non-linear dependencies among the labels~\cite{wicker2016auto} or develop a better feature space representation~\cite{read2015deep}. 
All these methods compress the label set regardless of the corresponding feature set. Recently, some feature-aware methods have been proposed that find the optimized compression function considering both feature and label sets~\cite{ChenL12,lin2014,li2015multi}. 

Although multi-label classification has received a lot of attention, classifying multi-label data streams is relatively new and not very well investigated. Some well-known methods like Binary Relevance (BR) and Classifier Chains (CC) can easily be upgraded to the online scenario by using an updateable classifier as their base learner. Qu et al.~\cite{qu2009mining} proposed a stacking version of Binary Relevance (MBR) with a weighting scheme in order to learn the dependencies among the labels via a meta-level classifier. Later, a multi-label version of Hoeffding Trees, a popular decision tree classifier in single-label stream mining, was presented~\cite{read2012scalable}. MLHT extends Hoeffding Trees to the multi-label scenario by modifying the definition of entropy as well as keeping multi-label classifiers at the leaves such as the majority-labelset classifier. Another method used a modified version of K-Nearest Neighbour, which is called Multiple Windows Classifier (MWC)~\cite{spyromitros2011dealing}. MWC maintains two fixed-size windows per label: one for positive and one for negative examples. Experimental comparisons show the dominance of the Multi-Label Hoeffding Tree over Meta Binary Relevance (MBR) and the Multiple Windows Classifier (MWC)~\cite{read2012scalable}. 
To the best of our knowledge, none of the available label space reduction methods in the literature on multi-label classification has been adapted for the streaming scenario yet. Most of them are computationally expensive, which makes them infeasible for streaming data.

To make our method efficient, we select the encoding matrix randomly. Choosing random features or projections has been an active topic in machine learning. Random features have been reasonably successful in scaling kernel methods when dealing with large data sets~\cite{rahimi2007random,le2013fastfood,dai2014scalable}. 
Training neural networks with random weights has also been elaborated in the literature~\cite{schmidt1992,widrow2013noprop}. 
 Studies show that there is significant redundancy in the parametrization of several deep learning methods and many of them even need not be learned at all~\cite{denil2013predicting}. Subsequent studies propose using random features to reduce the number of parameters for deep convolutional neural networks without sacrificing predictive performance~\cite{yang2015deep}. 
 Another study shows that features from a one-layer convolutional pooling architecture with completely random filters could achieve an average recognition rate that is just slightly worse than unsupervised pre-trained and fine-tuned filters~\cite{jarrett2009best}. 
  Further investigations show that a surprising fraction of the performance can be attributed to the architecture alone~\cite{saxe2011random}. 

\section{Online Label Compression Method} \label{method}
The key idea of our method is to compress labels into a smaller random space. As reviewed in the previous section, it has been shown that using random features or weights has been a successful component of many existing methods. Using a fixed random encoder will help in accelerating the method and make it scalable for streaming data. However, it may be too restrictive for multi-label stream classification. Therefore, we propose two variants of the compression function: one fixed and one adaptive encoder. 
 After receiving the first batch of data, we first map the label space to the reduced space. As the encoding is a real-valued mapping, the reduced labels will be real-valued. One may then use an updateable regressor or binarize the labels and train one updateable classifier for each pseudo label. The decoding function is calculated as a least squares solution and can be updated incrementally. The next batch of data is first used as test data: We obtain the prediction by the classifiers/regressors and use the decoding function to obtain predicted labels in the original space. After receiving the actual labels of the batch, the batch is used as training data to update the models and the decoding function accordingly. The remainder of this section will explain the algorithm in detail. 

\SetKwInput{KwInput}{Input}
\SetKwInput{KwOutput}{Output}

\begin{algorithm}[h!]
\caption{RACE Algorithm}\label{alg:mainPseudo}
\DontPrintSemicolon
\KwInput{$B_t = (X_{(t)},L_{(t)})$ is the $t^{th}$ batch of data,\\ 
\Indp \Indp $l$ and $k$ are the size of original and reduced label space respectively}
\KwOutput{encoder matrix $A_{l \times k}$, decoder matrix $\beta_{k \times l}$, and trained model}

\nl	Generate a random matrix $A_{0} = {(\mathbf{a_1}, \mathbf{a_2}, ..., \mathbf{a_k})}$ of size $l\times k$, where $\mathbf{a_i}$ is a vector of length $l$\;
	\tcp{Gram-Schmidt orthogonalization}
\nl \For{$i = 1$ to $k$}{
\nl		$\mathbf{v} = \mathbf{a_i}$\;
\nl		\For{$j = 1$ to $i-1$}{
\nl			$\mathbf{v} = \mathbf{v} - (\mathbf{a_j} \cdot \mathbf{v})\cdot \mathbf{a_j}$\;
		}	
\nl		$\mathbf{a_i} = \mathbf{v}/||\mathbf{v}||$
	}\:
	\tcp{Initialization step}
\nl	$H_0 = L_0A_0$\;
\nl \If{$Method$ = $classification$}{
\nl $H_0^{i,j} =
  \begin{cases}
   1 & \text{if } H_0^{i,j} \geq 0 \\
   0       & \text{otherwise}
  \end{cases}$ \;
  }
\nl Train a Binary Relevance updatable model on $(X_0,H_0)$\;
\nl Initialize label decoders:\;
	$K_0 = ({H_0}^{T}H_0)^{-1}$ , $\beta_0 = K_0H_0^{T}L_0$\;
	\tcp{Sequential step}
\nl \While{more batches of data}{
		\tcp{test new batch}
\nl		get $P_{t+1}$ the prediction of model on $X_{t+1}$\;
\nl		$Y_{t+1} = P_{t+1} \beta_t =
  		\begin{cases}
   			1 & \text{if } Y_{t+1}^{i,j} \geq 0 \\
   			0       & \text{otherwise}
  		\end{cases}$ \;
		\tcp{Update with new batch}
\nl		\If{$Encoding$ = $adaptive$}{
\nl			$A_{t+1} = {\beta_t}^{T}$
		}
\nl		Get pseudo labels by $H_{t+1} = L_{t+1}A_{t+1}$ \;
\nl 	\If{$Method$ = $classification$}{
\nl 		$H_{t+1}^{i,j} =
  				\begin{cases}
   					1 & \text{if } H_{t+1}^{i,j} \geq 0 \\
   					0       & \text{otherwise}
 				\end{cases}$ \;
  		}
\nl		Update Binary Relevance model with batch $(B_{t+1},H_{t+1})$\;
\nl		Update $K_{t+1}$ and $\beta_{t+1}$ using equation~\eqref{woodEq} \;
}
\end{algorithm}

\subsection{Problem Setting}
Let $\mathbf{x} \in \mathcal{X}$ denote an instance in the instance space $\mathcal{X} \subseteq \mathbb{R}^m$, and let $\mathbf{l} = (l_1,l_2,\ldots,l_l)$ be the set of relevant labels represented in a binary vector of length $l$. At time stamp $t$, we receive a new batch of labeled instances $(X(t),L(t)) = \{(\mathbf{x}_1^t,\mathbf{l}_1^t),\allowbreak (\mathbf{x}_2^t,\mathbf{l}_2^t),\ldots,(\mathbf{x}_n^t,\mathbf{l}_n^t)\}$ of size $n$, where $\mathbf{x}_i \in \mathcal{X}$ and $\mathbf{l}_i \in \{0,1\}^l$ denote the $i^{th}$ instance in the batch and its label set, respectively. 

A multi-label classifier $\mathcal{C}$ is a $\mathcal{X} \rightarrow \{0,1\}^l$ mapping that, for a given instance $\mathbf{x} \in \mathcal{X}$, returns a vector $\mathbf{y} = (y_1,y_2,\ldots,y_l)$ of predicted labels.
Our goal is to learn a predictor $\mathcal{C}: \mathcal{X} \rightarrow \{0,1\}^l$ that minimizes a label-wise decomposable loss function. To do so, we decompress the reduced label set by a decoding matrix ($\beta$) which minimizes the least squares error:
\begin{equation}
\argmin_{\beta_t} \sum_{i=1}^{n}{\mathbf{e}_t^2(i)},
\label{objective}
\end{equation}
where $\mathbf{e}_t(i)$ is the error of predicted labels for the $i^{th}$ instance of batch $t$. 
Linear models obtained by ordinary least squares are the same as for Hamming loss \cite{dembczynski2012}:

\begin{equation}
E_H(\mathbf{l},\mathbf{y}) = \frac{1}{l}\sum_{k=1}^{l}{\llbracket l_k \neq y_k \rrbracket}.
\label{hammingLoss}
\end{equation}

\subsection{Label Compression with Least Squares Solution}
Compressing the label space to a fixed random space was inspired by the idea of Extreme Learning Machines (ELMs)~\cite{huang2004extreme,liang2006fast}. 
Let $q = t . n$ be the total number of instances up to time $t$ and $L(q)$ denote the original label set, which is encoded into a smaller random space by $H(q) = L(q)A$, where $A$ is an $l \times k$ fixed encoder and $H(q)$ is the $q \times k$ resulting pseudo label matrix and $k$ is the reduced label space size. 
To decode the reduced label predictions to the original label space, we have:
\begin{equation}
H(q)\beta(q) = Y(q), 
\label{batchEq}
\end{equation}
where $\beta(q)$ is a $k \times l$ decoding matrix after observing $q$ instances. 
In order to find $\beta(q)$ in equation~\eqref{batchEq}, we use the least squares approach to make it faster than iterative optimization methods (such as stochastic gradient descent). 
Let $E(q) = [\mathbf{e}(1) \mathbf{e}(2) \ldots \mathbf{e}(q)]^T = L(q)-Y(q)$ be the error of predicted labels for all instances up to time $t$ according to the parameters at time $t$.  
By rewriting equation~\eqref{objective}, we obtain:
\begin{equation*}
\begin{split}
\xi &=  E^T(q)E(q) \\
	&= (L^T(q)-Y^T(q))(L(q)-Y(q)) \\
	&= (L^T(q)-\beta^T(q)H^T(q))(L(q)-H(q)\beta(q)) \\
	&= L^T(q)L(q)-2(H^T(q)L(q))^T\beta(q)+\beta^T(q)H^T(q)H(q)\beta(q).
\end{split}
\end{equation*}
Setting the gradient of $\xi$ with respect to $\beta(q)$ equal to zero, we obtain:
\begin{align}
(H^T(q)H(q))\hat{\beta}(q) = H^T(q)L(q), \nonumber \\
\hat{\beta}(q) = (H^T(q)H(q))^{-1}H^T(q)L(q). \label{lsSol}
\end{align}


Consequently, the resulting decoder matrix ($\hat{\beta}(q)$) is the least-squares solution of equation~\eqref{batchEq}. Experimental results show that this approach has better generalization performance at higher learning speed on both classification and regression problems \cite{huang2004extreme,farhang2013adaptive}.

\subsection{Random Compression of Multi-label Streams} \label{raceMethod}
The least-squares solution provided by equation~\eqref{lsSol} is of little interest in mining a stream of data, as it requires all the past samples at each iteration. We need to transform the solution in equation~\eqref{lsSol} into a sequential form. Hence, for the first batch of data, we calculate the decoding matrix as:
\begin{equation}
\beta_{0} = (H^T_{0} H_{0})^{-1}  H^T_{0} L_{0}. 
\label{beta0Eq}
\end{equation}
Suppose we are at step $t+1$ and receive a new batch of data. The new decoding matrix will be obtained by: 
\begin{equation}
\begin{split}
\beta_{t+1} &= 
\left(
\begin{bmatrix} H_{t} \\ H_{t+1}\end{bmatrix} ^{T}
\begin{bmatrix} H_{t} \\ H_{t+1}\end{bmatrix}
\right)
^{-1}
\begin{bmatrix} H_{t} \\ H_{t+1}\end{bmatrix}
^{T}
 \begin{bmatrix} L_{t} \\ L_{t+1}\end{bmatrix} \\
 &= 
 \left(H_{t}^{T}H_{t} + H_{t+1}^{T}H_{t+1} \right) ^{-1} 
 \left(H_{t}^{T}L_{t} + H_{t+1}^{T}L_{t+1} \right).
 \end{split}
 \label{beta1th}
 \end{equation}
 
To simplify equation~\eqref{beta1th}, we substitute equation~\eqref{batchEq} as an estimate of $L_{t}$. The second part of the right-hand side of equation~\eqref{beta1th} becomes:
\begin{equation} 
\begin{split}
\resizebox{.25\hsize}{!}{$H_{t}^{T}H_{t}\beta_{t} + H_{t+1}^{T}L_{t+1}$} &=\resizebox{.6\hsize}{!}{$\phantom{=}(H_{t}^{T}H_{t}+H_{t+1}^{T}H_{t+1}-H_{t+1}^{T}H_{t+1})\beta_{t} + H_{t+1}^{T}L_{t+1}$}\\
 &=\resizebox{.6\hsize}{!}{$\phantom{=}(H_{t}^{T}H_{t}+H_{t+1}^{T}H_{t+1})\beta_{t}-H_{t+1}^{T}H_{t+1}\beta_{t} + H_{t+1}^{T}L_{t+1}.$}\\
 \end{split}
\label{betaEq}
\end{equation}
 
Substituting equation~\eqref{betaEq} in \eqref{beta1th}, $\beta_{t+1}$ can be written as:  
\begin{equation} 
\beta_{t+1} = \beta_{t} + \left(H_{t}^{T}H_{t} + H_{t+1}^{T}H_{t+1} \right) ^{-1} H_{t+1}^{T}\left(L_{t+1} - H_{t+1}\beta_{t}\right).
\label{betaFin}
\end{equation}

In order to avoid multiple calculations of matrix inversions in equation~\eqref{betaFin}, we use the Sherman-Morrison-Woodbury formula, where rather than keeping $M_{t} = H_{t}^{T}H_{t}$, we keep $K_{t} = M_{t}^{-1}$. We can rewrite $\beta_{t+1}$ as the following: 
\begin{equation}
\begin{split}
M_{t+1}^{-1} &= (M_{t} + H_{t+1}^{T}H_{t+1})^{-1} \\
			 &= M_{t}^{-1} - M_{t}^{-1}H_{t+1}^{T}(I + H_{t+1}M_{t}^{-1}H_{t+1}^{T})^{-1}H_{t+1}M_{t}^{-1}\\
K_{t+1} &= K_{t} - K_{t}H_{t+1}^{T}(I+H_{t+1}K_{t}H_{t+1}^{T})^{-1}H_{t+1}K_{t}\\
\beta_{t+1} &= \beta_{t} + K_{t+1}H_{t+1}^{T}\left(L_{t+1} - H_{t+1}\beta_{t}\right).
\end{split}
\label{woodEq}
\end{equation}

The recursive method for updating the least-squares solution of $\beta_{t+1}$ is similar to the Recursive Least-Squares (RLS) algorithm \cite{farhang2013adaptive}. 
Algorithm~\ref{alg:mainPseudo} indicates the main steps of our proposed online label compression (\AlgName) method. The first step is to generate the encoding matrix (line $1$). We generate $k$ uniformly distributed random hyperplanes in the space of labels, where $k$ is the dimension of the reduced label space. In order to have more informative pseudo labels, we use the Gram-Schmidt algorithm \cite{trefethen1997numerical} to rotate the $k$ hyperplanes orthogonally. The Gram-Schmidt algorithm (line $2-6$) subtracts from vector $\mathbf{a_i}$ its components along previously determined orthonormal directions ${\mathbf{a_1},...,\mathbf{a_{(i-1)}}}$ to obtain the new orthogonal direction $\mathbf{v} = \mathbf{a_i} - \sum_{j=1}^{i-1} {(\mathbf{a_j} \times \mathbf{a_i})\mathbf{a_j}}$. Then, scale $\mathbf{v}$ to obtain a unit norm vector: $\mathbf{a}_i = \mathbf{v}/||\mathbf{v}||$. These orthonormal hyperplanes are kept as the encoding matrix or as an initialization of the adaptive variant.

For the first batch of instances, the \AlgName algorithm finds the pseudo labels by multiplying the label matrix with the encoding matrix (line $7$), and if the binary relevance models are chosen to be classifiers, the resulting real-valued pseudo labels are converted to binary values (line $8-9$). It then trains a binary relevance model, either a regressor or a classifier, on the encoded data (line $10$). Finally, the decoding matrix $\beta$ is initialized based on equation~\eqref{beta0Eq}. 
Lines $12-21$ repeat this procedure for the following batches of data. Before updating the models and the decoding matrix, we use the current model and $\beta$ to predict the labels of the newly arrived batch: The model predicts the labels in the reduced space (line $13$), and the decoding matrix of the previous step ($\beta_t$) transfers pseudo labels to the original space (line $14$).
To update the model with the new batch, we first update the encoding matrix to the transpose of the decoding matrix from the previous step for the adaptive encoding variants (line $15-16$). The pseudo labels of the new batch are extracted from its true labels (line $17-19$) and used to update the learning model (line $20$). In the end, $\beta$ is incrementally updated by equation~\eqref{woodEq}. 

\begin{table}
\begin{center}
\caption{Multi-label benchmark datasets used in experiments. $\mid$D$\mid$, $\mid$X$\mid$, $\mid$L$\mid$, LC, LD and UL indicate number of instances, number of features, number of labels, label cardinality, label density, and unique label sets respectively.} \label{tab:datasets}
{
\renewcommand{\arraystretch}{1.2}
\resizebox{0.8\textwidth}{!}{
\begin{tabular}{  @{}l | l | l | l | l | l | r } 
  \hline
  Name & $\mid$D$\mid$  & $\mid$X$\mid$  & $\mid$L$\mid$  & LC & LD & UL \\
  \hline 
  CAL500 & 502 & 68 & 174 & 26.044 & 0.150 & 502\\
  delicious & 16105 & 500 & 983 & 19.020 & 0.019 & 15806\\
  enron & 1702 & 1001 & 53 & 3.378 & 0.064 & 753\\ 
  mediamill & 43907 & 120 & 101 & 4.376 & 0.043 & 6555\\
  NUS-WIDE & 269648 & 500 & 81 & 1.869 & 0.023 & 18430\\
 rcv1v2(subset1) & 6000 & 47236 & 101 & 2.880 & 0.029 & 1028\\
  \hline
\end{tabular}
}
}
\end{center}
\end{table} 

\begin{table*}[tp]
\caption{Comparison of \AlgName variants with respect to different measures per dataset. Each cell indicates the mean and standard deviation of all runs including ($rank$).} \label{tab:raceVars} 
\centering
\resizebox{\textwidth}{!}{
\renewcommand\textfraction{.1}
\renewcommand{\floatpagefraction}{.1}%
\renewcommand{\topfraction}{.1}

\begin{tabular}[t]{ c|l|l|HHccHHHcHcHHHHcHHHH|c|r}
 Dataset&  Method & Encoding&Average Precision& Coverage &
Ex.-based accuracy &  Ex.-based F-measure &
Ex.-Based Precision & Ex.-Based Recall &
Ex.-Based Specificity &  Hamming loss &
Macro-avg. AUC &
Macro-avg. F-measure & Macro-avg. Precision &
Macro-avg. Recall &  Macro-avg. Specificity &
Micro-avg. AUC & Micro-avg. F-measure &
Micro-avg. Precision &  Micro-avg. Recall &
Micro-avg. Specificity &  Subset Accuracy &
\textbf{Average rank} &  Running time (s) \\ \hline \hline

\multirow{6}{*}{CAL500}   & Classification & Fixed & \textbf{0.462} & 143.633 & \textbf{0.22 $\pm$ 0.01} $(1)$ & \textbf{0.35 $\pm$ 0.02} $(1)$ & 0.509 & 0.279 & 0.951 & 0.15 $\pm$ 0.01 $(2)$ & 0.524 & \textbf{0.21 $\pm$ 0.01} $(1)$ & 0.223 & 0.208 & 0.921 & \textbf{0.776} & \textbf{0.35 $\pm$ 0.02} $(1)$ & 0.496 & 0.275 & 0.951 & 0.000 & 1.2 & 2.69\\ 
 &  & Adaptive & 0.408 & 155.287 & 0.21 $\pm$ 0.02 $(2)$ & 0.33 $\pm$ 0.02 $(2)$ & 0.380 & 0.389 & 0.827 & 0.24 $\pm$ 0.04 $(4)$ & 0.514 & 0.19 $\pm$ 0.02 $(2)$ & 0.190 & 0.249 & 0.800 & 0.664 & 0.33 $\pm$ 0.03 $(2)$ & 0.293 & 0.389 & 0.826 & 0.000 & 2.4 & 2.57\\ 
 \cdashline{2-24}
 & Regression & Fixed & 0.443 & 145.984 & 0.20 $\pm$ 0.01 $(3)$ & 0.33 $\pm$ 0.01 $(2)$ & 0.527 & 0.247 & 0.957 & 0.15 $\pm$ 0.00 $(2)$ & 0.507 & 0.17 $\pm$ 0.01 $(3)$ & 0.184 & 0.183 & 0.929 & 0.745 & 0.32 $\pm$ 0.01 $(3)$ & 0.506 & 0.240 & 0.957 & 0.000 & 2.6 & 4.29\\
 &  & Adaptive & 0.460 & 142.233 & 0.19 $\pm$ 0.01 $(4)$ & 0.31 $\pm$ 0.01 $(4)$ & 0.567 & 0.222 & 0.969 & \textbf{0.14 $\pm$ 0.00} $(1)$ & 0.514 & 0.16 $\pm$ 0.01 $(4)$ & 0.165 & 0.176 & 0.942 & 0.772 & 0.31 $\pm$ 0.01 $(4)$ & 0.561 & 0.216 & 0.969 & 0.000 & 3.4 & 4.20\\
\hline 

\multirow{6}{*}{delicious}  & Classification & Fixed & 0.206 & 726.093 & \textbf{0.10 $\pm$ 0.04} $(1)$ & \textbf{0.17 $\pm$ 0.06} $(1)$ & 0.300 & 0.135 & 0.990 & 0.03 $\pm$ 0.00 $(3)$ & 0.529 & 0.07 $\pm$ 0.02 $(3)$ & 0.076 & 0.077 & 0.988 & 0.743 & \textbf{0.17 $\pm$ 0.06} $(1)$ & 0.229 & 0.139 & 0.990 & 0.000 & 1.8 & 278.54\\
 &  & Adaptive & 0.199 & 700.892 & \textbf{0.10 $\pm$ 0.01} $(1)$ & 0.16 $\pm$ 0.02 $(2)$ & 0.268 & 0.161 & 0.966 & 0.05 $\pm$ 0.00 $(4)$ & 0.576 & 0.05 $\pm$ 0.01 $(4)$ & 0.047 & 0.076 & 0.964 & 0.755 & 0.11 $\pm$ 0.02 $(2)$ & 0.083 & 0.153 & 0.966 & 0.000 & 2.6 & 278.17\\ 
  \cdashline{2-24} 
 & Regression & Fixed & 0.227 & 658.909 & 0.01 $\pm$ 0.00 $(3)$ & 0.02 $\pm$ 0.01 $(4)$ & 0.111 & 0.008 & 1.000 & \textbf{0.02 $\pm$ 0.00} $(1)$ & 0.541 & \textbf{0.08 $\pm$ 0.00} $(1)$ & 0.087 & 0.083 & 1.000 & 0.809 & 0.02 $\pm$ 0.01 $(3)$ & 0.513 & 0.008 & 1.000 & 0.001 & 2.4 & 2343.54\\ 
 &  & Adaptive & 0.209 & 689.963 & 0.01 $\pm$ 0.00 $(3)$ & 0.03 $\pm$ 0.01 $(3)$ & 0.113 & 0.015 & 0.997 & \textbf{0.02 $\pm$ 0.00} $(1)$ & 0.540 & \textbf{0.08 $\pm$ 0.00} $(1)$ & 0.083 & 0.082 & 0.997 & 0.799 & 0.02 $\pm$ 0.00 $(3)$ & 0.454 & 0.014 & 0.997 & 0.001 & 2.2 & 2815.23\\ 
\hline 

 \multirow{6}{*}{enron}  & Classification & Fixed & 0.487 & 18.979 & 0.14 $\pm$ 0.07 $(4)$ & 0.21 $\pm$ 0.09$(4)$ & 0.417 & 0.239 & 0.889 & 0.15 $\pm$ 0.03 $(4)$ & 0.546 & 0.19 $\pm$ 0.08 $(4)$ & 0.087 & 0.125 & 0.884 & 0.812 & 0.18 $\pm$ 0.09 $(4)$ & 0.153 & 0.251 & 0.890 & 0.018 & 4 & 21.56 \\
 &  & Adaptive & 0.491 & 22.921 & \textbf{0.26 $\pm$ 0.03} $(1)$ & 0.35 $\pm$ 0.03 $(2)$ & 0.419 & 0.362 & 0.945 & 0.09 $\pm$ 0.00 $(3)$ & 0.568 & 0.21 $\pm$ 0.03 $(3)$ & 0.214 & 0.224 & 0.934 & 0.795 & 0.35 $\pm$ 0.04 $(2)$ & 0.339 & 0.371 & 0.945 & 0.033 & 2.2 & 20.39\\ 
  \cdashline{2-24}
 & Regression & Fixed & 0.543 & 18.161 & \textbf{0.26 $\pm$ 0.02} $(1)$ & \textbf{0.36 $\pm$ 0.03} $(1)$ & 0.523 & 0.299 & 0.984 & \textbf{0.06 $\pm$ 0.00} $(1)$ & 0.563 & \textbf{0.33 $\pm$ 0.01} $(1)$ & 0.336 & 0.331 & 0.971 & 0.840 & \textbf{0.38 $\pm$ 0.03} $(1)$ & 0.570 & 0.292 & 0.984 & 0.029 & 1 & 96.58\\ 
 &  & Adaptive & 0.560 & 17.353 & 0.23 $\pm$ 0.04 $(3)$ & 0.32 $\pm$ 0.06 $(3)$ & 0.527 & 0.254 & 0.989 & \textbf{0.06 $\pm$ 0.00} $(1)$ & 0.567 & 0.32 $\pm$ 0.01 $(2)$ & 0.329 & 0.322 & 0.979 & 0.848 & 0.34 $\pm$ 0.06 $(3)$ & 0.654 & 0.252 & 0.989 & 0.021 & 2.4 & 86.65\\ 
 \hline

\multirow{6}{*}{mediamill}  & Classification & Fixed & 0.514 & 44.567 & 0.22 $\pm$ 0.03 $(4)$ & 0.33 $\pm$ 0.03 $(4)$ & 0.305 & 0.527 & 0.906 & 0.11 $\pm$ 0.02 $(3)$ & 0.555 & 0.34 $\pm$ 0.05 $(3)$ & 0.339 & 0.408 & 0.893 & 0.767 & 0.30 $\pm$ 0.04 $(3)$ & 0.226 & 0.509 & 0.906 & 0.004 & 3.4 & 401.29\\ 
 &  & Adaptive & 0.546 & 39.609 & 0.25 $\pm$ 0.01 $(3)$ & 0.34 $\pm$ 0.02 $(3)$ & 0.389 & 0.453 & 0.857 & 0.16 $\pm$ 0.00 $(4)$ & 0.543 & 0.32 $\pm$ 0.00 $(4)$ & 0.057 & 0.118 & 0.848 & 0.778 & 0.22 $\pm$ 0.02 $(4)$ & 0.157 & 0.447 & 0.857 & 0.023 & 3.6 & 298.44\\ 
  \cdashline{2-24}
 & Regression & Fixed & \textbf{0.642} & 21.591 & \textbf{0.34 $\pm$ 0.01} $(1)$ & \textbf{0.45 $\pm$ 0.01} $(1)$ & 0.717 & 0.352 & 0.995 & \textbf{0.03 $\pm$ 0.00} $(1)$ & 0.553 & \textbf{0.45 $\pm$ 0.00} $(1)$ & 0.450 & 0.447 & 0.983 & \textbf{0.898} & \textbf{0.45 $\pm$ 0.01} $(1)$ & 0.735 & 0.324 & 0.995 & 0.050 & 1 & 504.65 \\ 
 &  & Adaptive & 0.635 & 21.628 & \textbf{0.34 $\pm$ 0.00} $(1)$ & \textbf{0.45 $\pm$ 0.00} $(1)$ & 0.709 & 0.350 & 0.994 & 0.04 $\pm$ 0.00 $(2)$ & 0.514 & 0.44 $\pm$ 0.00 $(2)$ & 0.445 & 0.447 & 0.981 & 0.884 & \textbf{0.45 $\pm$ 0.00} $(1)$ & 0.713 & 0.326 & 0.994 & 0.048 & 1.4 & 531.56\\
\hline 

\multirow{6}{*}{NUS-WIDE}  & Classification & Fixed & 0.363 & 25.759 & 0.17 $\pm$ 0.02 $(4)$ & 0.21 $\pm$ 0.02 $(4)$ & 0.213 & 0.304 & 0.938 & 0.08 $\pm$ 0.02 $(4)$ & 0.577 & 0.13 $\pm$ 0.08 $(4)$ & 0.035 & 0.100 & 0.935 & 0.703 & 0.17 $\pm$ 0.02 $(2)$ & 0.118 & 0.341 & 0.938 & 0.094 & 3.6 & 2157.70 \\
 &  & Adaptive & 0.434 & 16.757 & \textbf{0.23 $\pm$ 0.01} $(1)$ & \textbf{0.26 $\pm$ 0.01} $(1)$ & 0.259 & 0.285 & 0.986 & 0.03 $\pm$ 0.00 $(3)$ & 0.583 & 0.40 $\pm$ 0.02 $(3)$ & 0.397 & 0.419 & 0.983 & 0.841 & \textbf{0.23 $\pm$ 0.04} $(1)$ & 0.250 & 0.211 & 0.986 & 0.162 & 1.8 & 2629.84\\
  \cdashline{2-24}
 & Regression & Fixed & 0.438 & 14.635 & \textbf{0.23 $\pm$ 0.00} $(1)$ & 0.23 $\pm$ 0.00 $(3)$ & 0.238 & 0.227 & 1.000 & \textbf{0.02 $\pm$ 0.00} $(1)$ & 0.533 & \textbf{0.42 $\pm$ 0.00} $(1)$ & 0.423 & 0.418 & 1.000 & 0.833 & 0.02 $\pm$ 0.01 $(4)$ & 0.477 & 0.009 & 1.000 & 0.222 & 2 & 2699.95 \\
 &  & Adaptive & 0.417 & 14.753 & \textbf{0.23 $\pm$ 0.00} $(1)$ & 0.24 $\pm$ 0.00 $(2)$ & 0.254 & 0.233 & 1.000 & \textbf{0.02 $\pm$ 0.00} $(1)$ & 0.525 & \textbf{0.42 $\pm$ 0.00} $(1)$ & 0.427 & 0.419 & 1.000 & 0.832 & 0.04 $\pm$ 0.01 $(3)$ & 0.467 & 0.024 & 1.000 & 0.221 & 1.6 & 2415.93 \\
 \hline
 
 \multirow{6}{*}{rcv1v2}  & Classification & Fixed & 0.259 & 46.636 & \textbf{0.10 $\pm$ 0.03} $(1)$ & \textbf{0.16 $\pm$ 0.05} $(1)$ & 0.149 & 0.287 & 0.923 & 0.10 $\pm$ 0.04 $(3)$ & 0.545 & 0.04 $\pm$ 0.02 $(3)$ & 0.039 & 0.105 & 0.919 & 0.659 & \textbf{0.15 $\pm$ 0.04} $(1)$ & 0.104 & 0.293 & 0.922 & 0.011 & 1.8 & 22739.73 \\
 &  & Adaptive & 0.231 & 56.187 & 0.08 $\pm$ 0.02 $(2)$ & 0.12 $\pm$ 0.02 $(2)$ & 0.165 & 0.215 & 0.843 & 0.18 $\pm$ 0.03 $(4)$ & 0.592 & 0.04 $\pm$ 0.02 $(3)$ & 0.041 & 0.137 & 0.843 & 0.612 & 0.07 $\pm$ 0.02 $(2)$ & 0.047 & 0.228 & 0.842 & 0.008 & 2.6 & 22723.76 \\
  \cdashline{2-24}
 & Regression & Fixed & 0.246 & 33.572 & 0.01 $\pm$ 0.00 $(4)$ & 0.02 $\pm$ 0.01 $(4)$ & 0.031 & 0.010 & 1.000 & \textbf{0.03 $\pm$ 0.00} $(1)$ & 0.651 & \textbf{0.19 $\pm$ 0.00} $(1)$ & 0.197 & 0.193 & 1.000 & 0.798 & 0.03 $\pm$ 0.01 $(4)$ & 0.455 & 0.011 & 1.000 & 0.000 & 2.8 & 59603.26 \\
 &  & Adaptive & 0.251 & 35.899 & 0.04 $\pm$ 0.01 $(3)$ & 0.06 $\pm$ 0.02 $(3)$ & 0.067 & 0.025 & 0.999 & \textbf{0.03 $\pm$ 0.00} $(1)$ & 0.549 & \textbf{0.19 $\pm$ 0.00} $(1)$ & 0.195 & 0.195 & 0.998 & 0.772 & 0.06 $\pm$ 0.03 $(3)$ & 0.231 & 0.023 & 0.999 & 0.005 & 2.2 & 57300.81 \\
 
\specialrule{1pt}{0.5pt}{0.5pt}
 
  \multirow{6}{*}{\textbf{Average rank}}  & Classification & Fixed &  &  & 2.5 & 2.5 &  &  &  & 3.17 &  & 3 &  &  &  &  & 2 &  &  &  &  & \textbf{2.63} \\
 &  & Adaptive &  &  & 1.67 & 2 &  &  &  & 3.67 &  & 3.17 &  &  &  &  & 2.17 &  &  &  &  & \textbf{2.54} \\
  \cdashline{2-23}
 & Regression & Fixed &  &  & 2 & 2.5 &  &  &  & 1.17 &  & 1.33 &  &  &  &  & 2.67 &  &  &  &  & \textbf{1.93} \\
 &  & Adaptive &  &  & 2.67 & 2.67 &  &  &  & 1.17 &  & 1.83 &  &  &  &  & 2.83 &  &  &  &  & \textbf{2.23} \\
 \cline{1-23}

\end{tabular}
}
\end{table*}
 
\section{Experimental Results} \label{expr}
Experiments\footnote{The RACE source code is available at~\url{https://github.com/kramerlab/RACE}} were performed in a prequential manner, where each data batch is first treated as test data and then as training data. All methods were developed within the Mulan framework~\cite{mulan}, the experiments have been repeated ten times to reduce the effect of random parameters (e.g., $A$ in RACE or the number of chains in OECC), and the average values are reported.

\subsection{Experimental Setting}
\textbf{Benchmark datasets.} We evaluated our proposed method on several multi-label dataset benchmarks (Table~\ref{tab:datasets}). We have chosen these datasets in order to cover different types of multi-label datasets: datasets with high label density (e.g. CAL500), datasets with a lot of labels (e.g. delicious), datasets with a large feature space (e.g. rcv1v2), and datasets with a large sample size (e.g. mediamill and NUS-WIDE). 


\textbf{Baseline and comparison methods.} We have evaluated four variants of \AlgName, where the encoding may be fixed or adaptive, and the learning method may be regression or classification. We have developed an online version of Binary Relevance (OBR) and Ensemble of Classifier Chains (OECC),  two well-known multi-label methods, to compare with \AlgName. ECCs are an extension of CCs that reduce the chance of poorly ordered chains and create more scalable classifiers for batch learning~\cite{read2011classifier}. Our experimental results in the online setting are in line with previous findings in the batch setting so that only OECC results are reported in our experiments. 
 As many multi-label datasets have a very sparse label space, we implemented an always negative classifier (Negative) to see how well the classifier learns the available labels. In addition, we implemented the majority prediction (Majority) baseline method that takes the cardinality of the current data batch as a threshold for the classification of the following batch. Hence, if the cardinality of the current batch is $c$, the top $\lfloor c \rfloor$ labels are predicted as positive and the rest is predicted as negative.

\textbf{Benchmark measures.} Multi-label classification algorithms can be evaluated by a variety of measures~\cite{zhang2013review,sorower2010literature}. 
One category of these measures are called bipartition measures which calculate the average differences of the actual and the predicted labels over all examples (example-based measures) or measure them for each label separately (label-based measures). Hamming loss and Example-based accuracy are examples of example-based measures. Label-based metrics can be obtained in micro or macro modes. Conceptually speaking, macro-averaged metrics give an equal weight to each label while micro-averaged measures give an equal weight to each example. In this paper we report Example-based accuracy, the Example-based F-measure, Hamming loss, the Micro-averaged and Macro-averaged F-measure, and the running time of all methods. 

\textbf{Parameter Settings.} Na\"{\i}ve Bayes Updateable is used as a simple generative updateable base classifier for OBR, OECC, and \AlgName (classification variants). For PLST and regression variants of \AlgName, we used Stochastic Gradiant Descent (SGD) with the squared loss function and a learning rate of $10^{-4}$. The size of ensemble in OECC is set to 5, and the size of the reduced label space in \AlgName to $k = \lceil \log_2l \rceil$, where $l$ is the size of the original label space. The window size is set to 50 instances for CAL500, 100 for enron, and 500 for the rest. 

\subsection{Comparison of RACE Variants}
We first compare the four variants of \AlgName, where the encoding matrix can be fixed or adaptive, and the learning method is either regression or classification. The results are reported in Table~\ref{tab:raceVars}. The worst performance belongs to the classification variant with fixed encoding. This can be due to the very confined structure of the model. On the other hand, the regression variant with adaptive encoding does not perform well, possibly due to the overfitting of so many real-valued parameters. 
 The average ranks over all datasets indicate that the classification variant with adaptive encoding exhibits the best performance in terms of Example-based accuracy and Example-based F-measure and nearly the best in terms of Micro-averaged F-measure. Conversely, the regression variant with fixed encoding achieves the best results in terms of Hamming loss and Macro-averaged F-measure. However, its Example-based measures and Micro-averaged F-measure on some datasets (i.e. delicious and rcv1v2) are poor. 
Concerning running time, the classification variants are faster than their regression counterparts as their base learners are Naive Bayes Updateable, which is faster than the Stochastic Gradient Descent regression model.   

\begin{table}[!h]
\caption{Comparison of \AlgName to OBR, OECC, and the Negative and Majority baseline methods across different measures per dataset ($rank$). } \label{tab:reMeasuresTable}
\centering
\resizebox{0.95\textwidth}{!}{
\renewcommand\textfraction{.1}
\renewcommand{\floatpagefraction}{.2}%
\renewcommand{\topfraction}{.2}

\begin{tabular}[t]{ c|l|HHccHHHcHcHHHHcHHHH|c|r}
 \begin{turn}{-90} Dataset\end{turn}& \begin{turn}{-90} Method\end{turn}& \begin{turn}{-90} Average Precision\end{turn} & \begin{turn}{-90} Coverage\end{turn} &
\begin{turn}{-90} Ex.-based accuracy\end{turn} & \begin{turn}{-90} Ex.-based F-measure\end{turn} &
\begin{turn}{-90} Ex.-Based Precision\end{turn} & \begin{turn}{-90} Ex.-Based Recall\end{turn} &
\begin{turn}{-90} Ex.-Based Specificity\end{turn} & \begin{turn}{-90} Hamming loss\end{turn} &
\begin{turn}{-90} Macro-avg. AUC\end{turn} &
\begin{turn}{-90} Macro-avg. F-measure\end{turn} & \begin{turn}{-90} Macro-avg. Precision\end{turn} &
\begin{turn}{-90} Macro-avg. Recall\end{turn} & \begin{turn}{-90} Macro-avg. Specificity\end{turn} &
\begin{turn}{-90} Micro-avg. AUC\end{turn} & \begin{turn}{-90} Micro-avg. F-measure\end{turn} &
\begin{turn}{-90} Micro-avg. Precision\end{turn} & \begin{turn}{-90} Micro-avg. Recall\end{turn} &
\begin{turn}{-90} Micro-avg. Specificity\end{turn} & \begin{turn}{-90} Subset Accuracy\end{turn} &
\begin{turn}{-90} \textbf{Average rank}\end{turn} & \begin{turn}{-90} Running time (s)\end{turn}\\ \hline \hline

\multirow{6}{*}{CAL500}   & \AlgName (cls-adap) & 0.408 & 155.287 & 0.21 $(2)$ & 0.33 $(2)$ & 0.380 & 0.389 & 0.827 & 0.24 $(3)$ & 0.514 & 0.19 $(3)$ & 0.190 & 0.249 & 0.800 & 0.664 & 0.33 $(3)$ & 0.293 & 0.389 & 0.826 & 0.000 & 2.6 & 2.57\\ 
 & \AlgName (reg-fixed) & 0.443 & 145.984 & 0.20 $(4)$ & 0.33 $(2)$ & 0.527 & 0.247 & 0.957 & \textbf{0.15} $(1)$ & 0.507 & 0.17 $(4)$ & 0.184 & 0.183 & 0.929 & 0.745 & 0.32 $(4)$ & 0.506 & 0.240 & 0.957 & 0.000 & 3 & 4.29 \\
& Majority  & 0.120 & 171.089 & 0.03 $(5)$ & 0.06 $(5)$ & 0.063 & 0.062 & 0.837 & 0.28 $(5)$ & 0.500 & 0.11 $(6)$ & 0.103 & 0.216 & 0.852 & 0.500 & 0.06 $(5)$ & 0.063 & 0.062 & 0.837 & 0.000 & 5.2 & 9.79\\ 
 & Negative & 0.120 & 171.089 & 0.00 $(6)$ & 0.00 $(6)$ & 0.000 & 0.000 & 1.000 & \textbf{0.15} $(1)$ & 0.500 & 0.12 $(5)$ & 0.119 & 0.119 & 1.000 & 0.500 & 0.00 $(6)$ & 0.000 & 0.000 & 1.000 & 0.000 & 4.8 & 9.64\\ 
 & OBR  & 0.312 & 151.504 & \textbf{0.22} $(1)$ & \textbf{0.35} $(1)$ & 0.296 & 0.486 & 0.770 & 0.27 $(4)$ & 0.541 & \textbf{0.25} $(1)$ & 0.220 & 0.351 & 0.743 & 0.711 & \textbf{0.35} $(1)$ & 0.287 & 0.488 & 0.769 & 0.000 & 1.6 & 17.14\\ 
 & OECC  & 0.274 & 165.504 & 0.21 $(2)$ & 0.33 $(2)$ & 0.287 & 0.483 & 0.726 & 0.31 $(6)$ & 0.538 & 0.24 $(2)$ & 0.214 & 0.377 & 0.706 & 0.638 & 0.34 $(2)$ & 0.276 & 0.487 & 0.725 & 0.000 & 2.8 & 46.84\\ 
\hline 

\multirow{6}{*}{delicious}  & \AlgName (cls-adap) & 0.199 & 700.892 & \textbf{0.10} $(1)$ & 0.16 $(2)$ & 0.268 & 0.161 & 0.966 & 0.05 $(4)$ & 0.576 & 0.05 $(5)$ & 0.047 & 0.076 & 0.964 & 0.755 & 0.11 $(2)$ & 0.083 & 0.153 & 0.966 & 0.000 & 2.8 & 278.17\\
 & \AlgName (reg-fixed) & 0.227 & 658.909 & 0.01 $(4)$ & 0.02 $(4)$ & 0.111 & 0.008 & 1.000 & \textbf{0.02} $(1)$ & 0.541 & \textbf{0.08} $(1)$ & 0.087 & 0.083 & 1.000 & 0.809 & 0.02 $(4)$ & 0.513 & 0.008 & 1.000 & 0.001 & 2.8 & 2343.54\\ 
  & Majority & 0.029 & 899.116 & 0.01 $(4)$ & 0.01 $(5)$ & 0.009 & 0.011 & 0.981 & 0.04 $(3)$ & 0.500 & \textbf{0.08} $(1)$ & 0.081 & 0.099 & 0.981 & 0.500 & 0.01 $(5)$ & 0.009 & 0.009 & 0.981 & 0.000 & 3.6 & 11760.25\\ 
 & Negative & 0.029 & 899.116 & 0.00 $(6)$ & 0.00 $(6)$ & 0.001 & 0.001 & 1.000 & \textbf{0.02} $(1)$ & 0.500 & \textbf{0.08} $(1)$ & 0.083 & 0.083 & 1.000 & 0.500 & 0.00 $(6)$ & 0.000 & 0.000 & 1.000 & 0.001 & 4 & 11324.38\\ 
 & OBR & \textbf{0.254} & 661.196 & \textbf{0.10} $(1)$ & \textbf{0.17} $(1)$ & 0.122 & 0.621 & 0.840 & 0.16 $(5)$ & \textbf{0.692} & 0.07 $(4)$ & 0.049 & 0.315 & 0.835 & \textbf{0.812} & \textbf{0.13} $(1)$ & 0.073 & 0.620 & 0.840 & 0.000 & 2.4 & 22117.97\\ 
 & OECC  & 0.061 & 915.472 & 0.04 $(3)$ & 0.07 $(3)$ & 0.058 & 0.458 & 0.394 & 0.61 $(6)$ & 0.566 & 0.04 $(6)$ & 0.042 & 0.603 & 0.399 & 0.422 & 0.04 $(3)$ & 0.021 & 0.441 & 0.394 & 0.000 & 4.2 & 46525.20\\ 
\hline 

 \multirow{6}{*}{enron}  &  \AlgName (cls-adap) & 0.491 & 22.921 & \textbf{0.26} $(1)$ & 0.35 $(2)$ & 0.419 & 0.362 & 0.945 & 0.09 $(3)$ & 0.568 & 0.21 $(4)$ & 0.214 & 0.224 & 0.934 & 0.795 & 0.35 $(2)$ & 0.339 & 0.371 & 0.945 & 0.033 & 2.4 & 20.39 \\
 & \AlgName (reg-fixed) & 0.543 & 18.161 & \textbf{0.26} $(1)$ & \textbf{0.36} $(1)$ & 0.523 & 0.299 & 0.984 & \textbf{0.06} $(1)$ & 0.563 & \textbf{0.33} $(1)$ & 0.336 & 0.331 & 0.971 & 0.840 & \textbf{0.38} $(1)$ & 0.570 & 0.292 & 0.984 & 0.029 & 1 & 96.58 \\
  & Majority & 0.082 & 43.613 & 0.07 $(5)$ & 0.11 $(5)$ & 0.138 & 0.100 & 0.950 & 0.11 $(4)$ & 0.500 & 0.27 $(3)$ & 0.271 & 0.291 & 0.946 & 0.500 & 0.12 $(5)$ & 0.138 & 0.113 & 0.950 & 0.000 & 4.4 & 56.52 \\
 & Negative & 0.082 & 43.613 & 0.00 $(6)$ & 0.00 $(6)$ & 0.000 & 0.000 & 1.000 & 0.07 $(2)$ & 0.500 & 0.29 $(2)$ & 0.289 & 0.289 & 1.000 & 0.500 & 0.00 $(6)$ & 0.000 & 0.000 & 1.000 & 0.000 & 4.4 & 57.51 \\
 & OBR & 0.376 & 29.436 & 0.23 $(3)$ & 0.33 $(4)$ & 0.311 & 0.571 & 0.834 & 0.18 $(5)$ & 0.642 & 0.14 $(5)$ & 0.127 & 0.266 & 0.823 & 0.787 & 0.29 $(3)$ & 0.205 & 0.558 & 0.835 & 0.024 & 4 & 108.12 \\
 & OECC & 0.307 & 32.310 & 0.23 $(3)$ & 0.34 $(3)$ & 0.315 & 0.568 & 0.833 & 0.18 $(5)$ & 0.577 & 0.13 $(6)$ & 0.120 & 0.256 & 0.823 & 0.703 & 0.29 $(3)$ & 0.204 & 0.557 & 0.834 & 0.024 & 4 & 275.51 \\
 \hline

\multirow{6}{*}{mediamill}   &  \AlgName (cls-adap) & 0.546 & 39.609 & 0.25 $(2)$ & 0.34 $(2)$ & 0.389 & 0.453 & 0.857 & 0.16 $(4)$ & 0.543 & 0.32 $(4)$ & 0.057 & 0.118 & 0.848 & 0.778 & 0.22 $(2)$ & 0.157 & 0.447 & 0.857 & 0.023 & 2.8 & 298.44\\ 
 & \AlgName (reg-fixed) & \textbf{0.642} & 21.591 & \textbf{0.34} $(1)$ & \textbf{0.45} $(1)$ & 0.717 & 0.352 & 0.995 & \textbf{0.03} $(1)$ & 0.553 & \textbf{0.45} $(1)$ & 0.450 & 0.447 & 0.983 & \textbf{0.898} & \textbf{0.45} $(1)$ & 0.735 & 0.324 & 0.995 & 0.050 & 1 & 504.65 \\ 
  & Majority  & 0.080 & 67.489 & 0.07 $(5)$ & 0.13 $(5)$ & 0.137 & 0.133 & 0.966 & 0.07 $(3)$ & 0.500 & 0.42 $(3)$ & 0.421 & 0.444 & 0.962 & 0.500 & 0.13 $(5)$ & 0.137 & 0.127 & 0.966 & 0.000 & 4.2 & 450.23\\ 
 & Negative & 0.080 & 67.489 & 0.04 $(6)$ & 0.04 $(6)$ & 0.039 & 0.039 & 1.000 & 0.04 $(2)$ & 0.500 & 0.43 $(2)$ & 0.426 & 0.426 & 1.000 & 0.500 & 0.00 $(6)$ & 0.000 & 0.000 & 1.000 & 0.039 & 4.4 & 444.24\\
 & OBR  & 0.202 & 42.752 & 0.10 $(3)$ & 0.17 $(3)$ & 0.099 & 0.683 & 0.699 & 0.30 $(5)$ & \textbf{0.676} & 0.15 $(5)$ & 0.131 & 0.424 & 0.693 & 0.780 & 0.17 $(3)$ & 0.098 & 0.707 & 0.700 & 0.000 & 3.8 & 1258.34\\ 
 & OECC  & 0.167 & 52.005 & 0.08 $(4)$ & 0.15 $(4)$ & 0.088 & 0.659 & 0.653 & 0.35 $(6)$ & 0.639 & 0.15 $(5)$ & 0.130 & 0.429 & 0.650 & 0.688 & 0.15 $(4)$ & 0.085 & 0.688 & 0.654 & 0.000 & 4.6 & 3295.56\\ 
\hline 

\multirow{6}{*}{NUS-WIDE}  &  \AlgName (cls-adap) & 0.434 & 16.757 & \textbf{0.23} $(1)$ & \textbf{0.26} $(1)$ & 0.259 & 0.285 & 0.986 & 0.03 $(3)$ & 0.583 & 0.40 $(4)$ & 0.397 & 0.419 & 0.983 & 0.841 & \textbf{0.23} $(1)$ & 0.250 & 0.211 & 0.986 & 0.162 & 2 & 2629.84\\
 & \AlgName (reg-fixed) & 0.438 & 14.635 & \textbf{0.23} $(1)$ & 0.23 $(2)$ & 0.238 & 0.227 & 1.000 & \textbf{0.02} $(1)$ & 0.533 & \textbf{0.42} $(1)$ & 0.423 & 0.418 & 1.000 & 0.833 & 0.02 $(4)$ & 0.477 & 0.009 & 1.000 & 0.222 & 1.8 & 2699.95 \\
  & Majority & 0.071 & 42.278 & 0.02 $(6)$ & 0.02 $(6)$ & 0.021 & 0.018 & 0.983 & 0.04 $(4)$ & 0.500 & 0.41 $(3)$ & 0.406 & 0.411 & 0.983 & 0.500 & 0.01 $(5)$ & 0.005 & 0.005 & 0.983 & 0.015 & 4.8 & 5338.37 \\
 & Negative & 0.071 & 42.278 & 0.22 $(3)$ & 0.22 $(3)$ & 0.224 & 0.224 & 1.000 & \textbf{0.02} $(1)$ & 0.500 & \textbf{0.42} $(1)$ & 0.417 & 0.417 & 1.000 & 0.500 & 0.00 $(6)$ & 0.000 & 0.000 & 1.000 & 0.224 & 2.8 & 6506.92 \\
 & OBR & 0.204 & 24.329 & 0.06 $(4)$ & 0.10 $(5)$ & 0.063 & 0.504 & 0.745 & 0.26 $(6)$ & 0.712 & 0.08 $(5)$ & 0.058 & 0.376 & 0.744 & 0.768 & 0.11 $(2)$ & 0.059 & 0.652 & 0.746 & 0.001 & 4.4 & 11481.38\\
 & OECC & 0.178 & 26.807 & 0.06 $(4)$ & 0.11 $(4)$ & 0.066 & 0.471 & 0.767 & 0.24 $(5)$ & 0.673 & 0.08 $(5)$ & 0.059 & 0.361 & 0.766 & 0.713 & 0.11 $(2)$ & 0.061 & 0.622 & 0.767 & 0.002 & 4 & 34511.24 \\
 \hline
 
 \multirow{6}{*}{rcv1v2}  & \AlgName (cls-adap) & 0.231 & 56.187 & 0.08 $(3)$ & 0.12 $(3)$ & 0.165 & 0.215 & 0.843 & 0.18 $(4)$ & 0.592 & 0.04 $(6)$ & 0.041 & 0.137 & 0.843 & 0.612 & 0.07 $(3)$ & 0.047 & 0.228 & 0.842 & 0.008 & 3.8 & 22723.76 \\
 & \AlgName (reg-fixed) & 0.246 & 33.572 & 0.01 $(4)$ & 0.02 $(4)$ & 0.031 & 0.010 & 1.000 & \textbf{0.03} $(1)$ & 0.651 & \textbf{0.19} $(1)$ & 0.197 & 0.193 & 1.000 & 0.798 & 0.03 $(4)$ & 0.455 & 0.011 & 1.000 & 0.000 & 2.8 & 59603.26 \\
  & Majority & 0.113 & 53.158 & 0.00 $(5)$ & 0.00 $(5)$ & 0.004 & 0.004 & 0.979 & 0.05 $(3)$ & 0.500 & 0.18 $(3)$ & 0.178 & 0.185 & 0.979 & 0.500 & 0.00 $(5)$ & 0.004 & 0.003 & 0.979 & 0.000 & 4.2 & 24487.15 \\
 & Negative & 0.113 & 53.158 & 0.00 $(5)$ & 0.00 $(5)$ & 0.000 & 0.000 & 1.000 & \textbf{0.03} $(1)$ & 0.500 & \textbf{0.19} $(1)$ & 0.193 & 0.193 & 1.000 & 0.500 & 0.00 $(5)$ & 0.000 & 0.000 & 1.000 & 0.000 & 3.4 & 19761.63 \\
 & OBR & 0.229 & 38.463 & \textbf{0.13} $(1)$ & \textbf{0.20} $(1)$ & 0.157 & 0.860 & 0.560 & 0.43 $(5)$ & 0.627 & 0.13 $(4)$ & 0.098 & 0.626 & 0.560 & 0.703 & \textbf{0.12} $(1)$ & 0.067 & 0.834 & 0.560 & 0.022 & 2.4 & 234679.01 \\
 & OECC & 0.255 & 36.030 & \textbf{0.13} $(1)$ & 0.19 $(2)$ & 0.154 & 0.867 & 0.534 & 0.46 $(6)$ & 0.630 & 0.13 $(4)$ & 0.100 & 0.633 & 0.534 & 0.714 & \textbf{0.12} $(1)$ & 0.075 & 0.844 & 0.534 & 0.022 & 2.8 & 154585.83 \\

\specialrule{1pt}{0.5pt}{0.5pt}

 \multirow{6}{*}{\textbf{Average rank}}  & \AlgName (cls-adap) &  &  & 1.67 & 2 &  &  &  & 3.5 &  & 4.33 &  &  &  &  & 2.17 &  &  &  &  & \textbf{2.73} \\
 & \AlgName (reg-fixed) &  &  & 2.5 & 2.33 &  &  &  & 1 &  & 1.5 &  &  &  &  & 3 &  &  &  &  & \textbf{2.07} \\
  & Majority &  &  & 5 & 5.17 &  &  &  & 3.67 &  & 3.17 &  &  &  &  & 5 &  &  &  &  & \textbf{4.4} \\
 & Negative &  &  & 5.33 & 5.33 &  &  &  & 1.33 &  & 2 &  &  &  &  & 5.83 &  &  &  &  & \textbf{3.96} \\
 & OBR &  &  & 2.17 & 2.5 &  &  &  & 5 &  & 4 &  &  &  &  & 1.83 &  &  &  &  & \textbf{3.1} \\
 & OECC &  &  & 2.83 & 3 &  &  &  & 5.67 &  & 4.67 &  &  &  &  & 2.5 &  &  &  &  & \textbf{3.73} \\
 \cline{1-22}

\end{tabular}
}
\end{table}

\subsection{Comparison to Online Baseline Methods}
In Table \ref{tab:reMeasuresTable} we compare two variants of \AlgName , the classification method with adaptive encoding (cls-adap) and the regression method with fixed encoding (reg-fixed), to other online baseline methods and report the average values for different measures over all batches and runs. We overlooked the standard deviation of different runs in this table as the corresponding values for OECC were neglectable and the ones for the RACE variants were reported in Table~\ref{tab:raceVars}. 
 Again, the average ranks over all datasets indicate that  \AlgName (cls-adap) achieves the best Example-based accuracy and Example-based F-measure and nearly the best Micro-averaged F-measure; and hence, with the average rank of $2.73$ over all measures, it stands on the second place, after \AlgName (reg-fixed). \AlgName (reg-fixed) achieves the best results in terms of Hamming loss and Macro-averaged F-measure. It gives the best performance across all evaluated measures on enron and mediamill. However, its poor behavior in terms of Example-based measures and the Micro-averaged F-measure on delicious and rcv1v2 is quite similar to the Negative baseline. 

\begin{figure}
\includegraphics[width=0.8\linewidth]{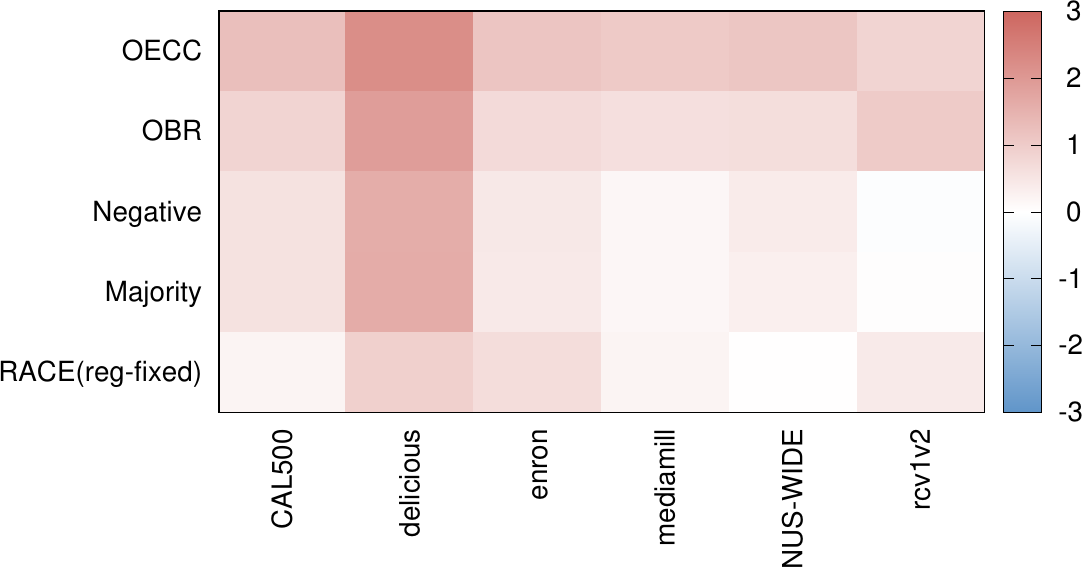}
\caption{Runtime comparison of different algorithms to RACE (cls-adap), results shown as log ratio.} \label{fig:timeHeatmap}
\end{figure}

The last column of Table \ref{tab:reMeasuresTable} presents the running time of all algorithms. For a better visualization of time complexity, we plotted a heat map that represents the log ratio of each method's time, i.e. $\log_{10}{\frac{t_{method}}{t_{RACE(cls-adap)}}}$ (Figure~\ref{fig:timeHeatmap}). The heat map illustrates the difference in the order of the time needed to finish for each method in comparison to \AlgName (cls-adap) on each dataset. We observe that the space reduction method is efficient and the running time of \AlgName (cls-adap) is orders of magnitude smaller than the one of other ensemble methods, especially when the original label space is very large (see the results for delicious).

\begin{table*}[tb]
\caption{Comparison of  \AlgName (cls-adap), its iterative version and PLST on various datasets. } \label{tab:offlineTable}
\centering
\resizebox{0.9\textwidth}{!}{
\begin{tabular}[t]{ c|*{5}{ccHc|}ccHc}  
 & \multicolumn{4}{c|}{CAL500} & \multicolumn{4}{c|}{delicious} & \multicolumn{4}{c|}{enron} & \multicolumn{4}{c|}{mediamill} & \multicolumn{4}{c|}{NUS-WIDE} & \multicolumn{4}{c}{rcv1v2}\\
 \cline{2-25}
 & \AlgName & \AlgName & Offline & PLST  & \AlgName & \AlgName & Offline & PLST  & \AlgName & \AlgName & Offline & PLST  & \AlgName & \AlgName & Offline & PLST  & \AlgName & \AlgName & Offline & PLST  & \AlgName & \AlgName & Offline & PLST \\
&  & ($iter=3$) &  &  &  & ($iter=3$) &  &  &  & ($iter=3$) &  &  &  & ($iter=3$) &  &  &  & ($iter=3$) &  &  &  & ($iter=3$) &  & \\ 
\hline
 Ex.-based accuracy & 0.23 $\pm$ 0.01 & \textbf{0.25 $\pm$ 0.01} & 0.23 & 0.19 & 0.10 $\pm$ 0.01 & \textbf{0.12 $\pm$ 0.04} & 0.07 & 0.03 & 0.33$\pm$ 0.04 & 0.33 $\pm$ 0.04 &  & \textbf{0.34} & 0.31 $\pm$ 0.02 & 0.29 $\pm$ 0.03 & 0.24 & \textbf{0.37} & \textbf{0.17 $\pm$ 0.03} & 0.15 $\pm$ 0.02 & & - & 0.10 $\pm$ 0.02 & \textbf{0.12 $\pm$ 0.02} &  & - \\
 Hamming loss & 0.17 $\pm$ 0.01 & 0.16 $\pm$ 0.01 & 0.15 & \textbf{0.14} & 0.04 $\pm$ 0.00 & 0.04 $\pm$ 0.01 & 0.04 & \textbf{0.02} & 0.07 $\pm$ 0.00 & 0.08 $\pm$ 0.04 & 0.08 & \textbf{0.06} & 0.06 $\pm$ 0.01 & 0.06 $\pm$ 0.00 & 0.11 & \textbf{0.03} & \textbf{0.04 $\pm$ 0.00} & \textbf{0.04  $\pm$ 0.00} & & - & \textbf{0.15 $\pm$ 0.04} & 0.19 $\pm$ 0.09 &  & - \\
 Running time (s) &  \textbf{1.79} & 4.46 & 6.98 & 2.34 & $\mathbf{1.43\times10^3}$  & $3.08\times10^3$ & $1.18\times 10^4$ & $2.28\times 10^4$ & \textbf{20.20} & 50.82 & 302.85 & 640.07 & $\mathbf{1.22\times 10^3}$ & $3.13\times 10^3$ & $1.08\times 10^5$ & $1.52\times 10^4$ & $\mathbf{6.97\times 10^4}$ & $1.62\times 10^5$ & & - & $\mathbf{1.84\times 10^4}$ & $2.28\times 10^4$ &  & - \\
\end{tabular}
}
\end{table*}

\subsection{Comparison to the Offline Label Compression Methods}
In this section we compare \AlgName (cls-adap) to PLST\footnote{We have used the implementation provided by the Meka framework at \url{https://github.com/Waikato/meka/tree/master/src/main/java/meka/classifiers/multilabel}.}~\cite{tai2012}, a popular offline label compression method which uses a projection method based on singular value decomposition. Here, we used hold-out evaluation, i.e. 33\% of each dataset was chosen randomly as the test set and the rest as the training set. \AlgName (cls-adap) received training data in batches and after updating for each batch, the test was performed on the test data, and the average values of Example-based accuracy and Hamming loss are reported. In addition to present each batch once to \AlgName , we repeated the experiment by showing every batch several times consecutively. Table~\ref{tab:offlineTable} presents the results for different measures and datasets, and reports the iterative results for $3$ iterations. We observe that while \AlgName has an adaptive nature, Example-based accuracy and Hamming loss are comparable to PLST and on some datasets (CAL500 and delicious) \AlgName reaches an even higher accuracy. Moreover, on all datasets, due to its random compression nature, \AlgName has an order of magnitude smaller running times, and when the dataset is large (e.g., rcv1v2 and NUS-WIDE), PLST does not even finish within reasonable time\footnote{The experiment was not finished after 120 hours on the same system.}. Looking into iterative \AlgName , we observe that this method achieves more stable results over time, however, on some datasets it leads to better accuracy (CAL500 and delicious), while on some others the accuracy is reduced (mediamill, NUS-WIDE, and rcv1v2). 

\begin{figure}[tb]
\includegraphics[width=\textwidth]{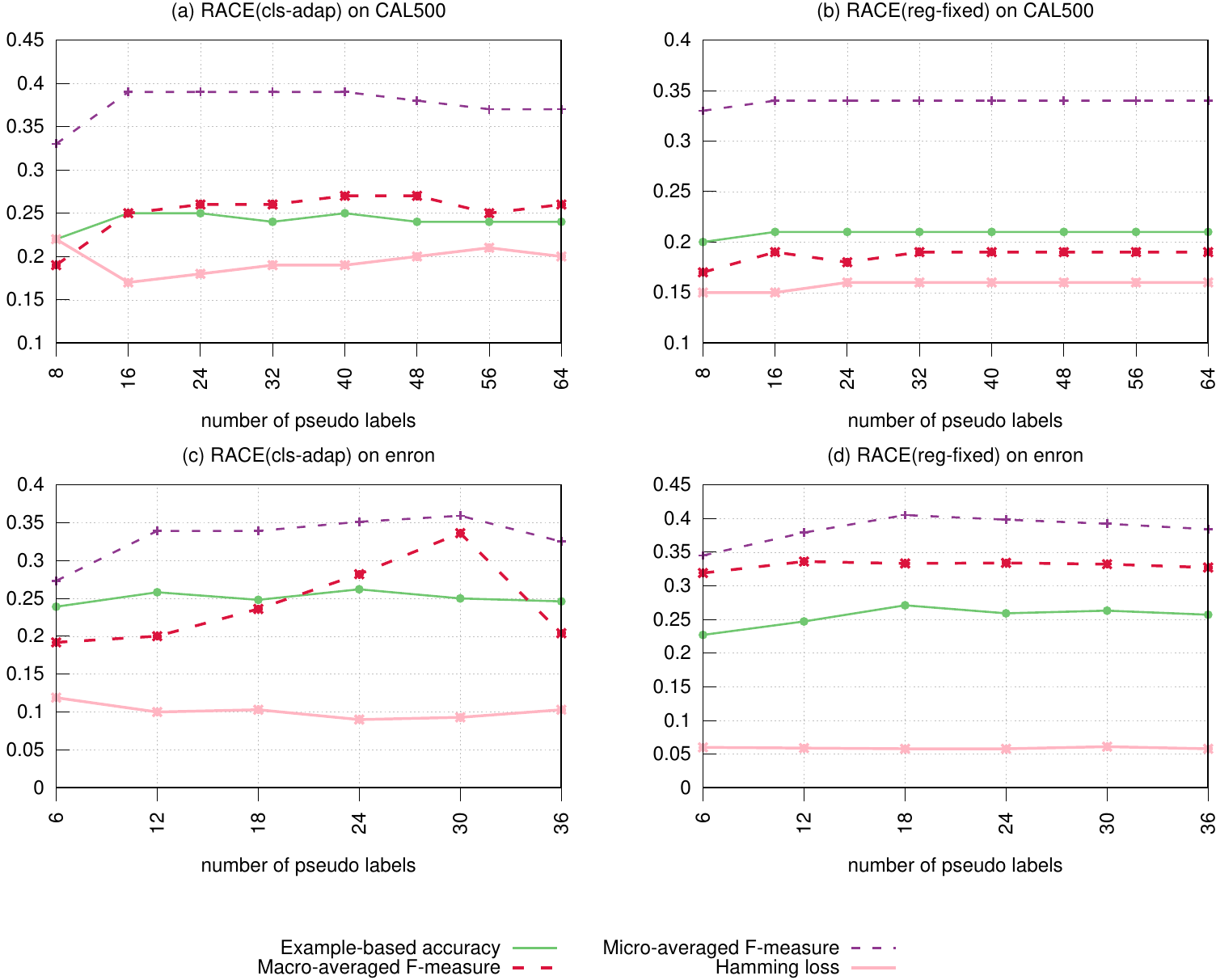}
\caption{Effect of different numbers of pseudo labels on the performance of \AlgName variants on the CAL500 and enron datasets.} \label{fig:pseudoEnron}
\end{figure}

\subsection{Impact of Pseudo Label Set Size} \label{labelSizeEffect}
\AlgName has one parameter to set: the size of the reduced label space. 
We have changed the number of pseudo labels from $\lceil \log_2l \rceil$ to ${\lceil \log_2l \rceil}^2$, where $l$ is the size of the original label space. Figure~\ref{fig:pseudoEnron} shows the impact of this parameter on the performance of \AlgName(cls-adap) and \AlgName(reg-fixed) for the CAL500 and the enron datasets. As we can see in both datasets, while increasing the number of pseudo labels in \AlgName(cls-adap) does not change Example-based accuracy and Hamming loss notably, the Micro-averaged and Macro-averaged F-measure are improved up to some point (at $30$ pseudo labels in enron and at $40$ pseudo labels in CAL500), but then they drop again. This is the case for \AlgName(reg-fixed) for Example-based accuracy and the Micro-averaged F-measure in the enron dataset, however, in the CAL500 dataset, different measures do not change notably for \AlgName(reg-fixed). Comparing these improvements to the average ranks from Table~\ref{tab:reMeasuresTable}, we can see that each of these variants can be improved by choosing a properly fine-tuned reduced label space size. 

\section{Conclusion} \label{conc}
This paper addresses the problem of online label compression for multi-label data stream classification. The main contribution of this paper is to encode the original label space by a random projection method to a much smaller space and decode the output of models by an incremental analytical approach inspired by Extreme Learning Machines. Different variants of the proposed method, \AlgName , were tested. The experimental evaluations showed that the approach works well on different datasets across a variety of measures, and outperforms other existing online multi-label baselines in terms of accuracy, F-measure, Hamming loss and notably running time. However, there is still room for further improvement. As future research, we plan to extend the current framework to cover various types of concept drift. An ensemble of \AlgName models also can be used to reduce the variance of its random initialization. 

\bibliographystyle{model2-names}
\bibliography{\jobname}

\end{document}